\pdfoutput=1

\documentclass[11pt]{article}

\usepackage{emnlp2021}

\usepackage{times}
\usepackage{latexsym}

\usepackage[T1]{fontenc}

\usepackage[utf8]{inputenc}

\usepackage{microtype}

\usepackage{tabu}
\usepackage{booktabs}
\usepackage{graphicx}
\usepackage{multicol}
\usepackage{multirow}
\usepackage{algorithm}
\usepackage[noend]{algpseudocode}
\usepackage{amsthm}
    \newtheorem{example}{Example}
\usepackage{amsmath}
\usepackage{enumerate}
\usepackage{enumitem}

\usepackage{xcolor}

\title{Automatic Rule Generation for Time Expression Normalization}

\author{Wentao Ding, Jianhao Chen, Jinmao Li, Yuzhong Qu \\
  State Key Laboratory for Novel Software Technology, Nanjing University, China\\
  \texttt{\{wtding, jhchen, jmli\}@smail.nju.edu.cn, yzqu@nju.edu.cn}}

\begin{document}
\maketitle
\begin{abstract}
The understanding of time expressions includes two sub-tasks: recognition and normalization. In recent years, significant progress has been made in the recognition of time expressions while research on normalization has lagged behind. Existing SOTA normalization methods highly rely on rules or grammars designed by experts, which limits their performance on emerging corpora, such as social media texts. In this paper, we model time expression normalization as a sequence of operations to construct the normalized temporal value, and we present a novel method called ARTime, which can automatically generate normalization rules from training data without expert interventions. Specifically, ARTime automatically captures possible operation sequences from annotated data and generates normalization rules on time expressions with common surface forms. The experimental results show that ARTime can significantly surpass SOTA methods on the Tweets benchmark, and achieves competitive results with existing expert-engineered rule methods on the TempEval-3 benchmark.
\end{abstract}

\section{Introduction}
Temporal information plays an important role in natural language. The research community divides the understanding of time expressions into two sub-tasks: recognition and normalization \citep{TEval3}. The first task is to annotate time expressions from free text, and the second one is to annotate the temporal values and types of the recognized time expressions. Some recent research work \citep{SynTime,TOMN,PTime} achieved significant improvements on the recognition task comparing with classic rule-based or semantic parsing systems, while the researches on normalization have lagged behind. Normalization methods often rely on expert-designed rules or grammars to model the compositional structure of time expression, which are domain-sensitive and not sufficient enough on covering emerging corpora.

To avoid the performance limitation and the labor cost of manually designing rules for different corpora, we study the problem of automatically generating normalization rules from annotated data. There are some challenges to achieve this goal. Firstly, the surface text forms of natural language expressions are diverse, and the normalized value of time expressions may not directly correspond to their surface text form. (e.g., both the expression ``May'' and ``this month'' could be normalized to ``2021-05''.)  Secondly, time expressions have rich semantic structures which are not explicitly reflected in their annotations. The implicitness of semantic structure makes supervised approaches hard to apply to the task of generating normalization rules. Besides, the annotations in practical datasets are noisy, which challenges the robustness of data-driven methods.

To achieve the goal, we regard time expression normalization as a sequence of operations to construct the normalized temporal value of specific types. We assume that the surface form of time expressions activates the corresponding normalization sequence. The normalization rules are defined as the alignment between surface form pattern and activated operation sequences, as demonstrated in Example \ref{eg:rule}. Section \ref{sec:def} will describe operations and normalization rules in details.

\begin{example}\label{eg:rule}
The time expression ``last October'' can be normalized by the rule (Pattern=``last MONTH:\$1'', Type=Instant, Operations=(\textsf{ToLast[Year], ModifyEnum[\$1]))}, where the type ``Instant'' indicates that the normalized value should be a date or time instant, the first operation decreases the current value on year field by $1$, and the second operation modifies the value on month field by the ``MONTH'' variable obtained from the expression (i.e., October).
\end{example}

We name the method for automatically generating normalization rules as ARTime \footnote{Our codes are available at \url{https://github.com/nju-websoft/ARTime}}. ARTime computes the difference between the base value and the annotated value of input time expression to capture possible operation sequences, aligning the captured sequence with the surface form of the time expression to construct candidate rules. It ranks the noisy candidates by their frequency to distinguish the good rules. When applying the rules for normalization, ARTime attempts to dynamically search a rule composition for unmatched expressions to improve the coverage of generated rules. The whole normalization process only relies on a small set of pre-defined lexicon of temporal values (e.g, numeric values and time units), and does not need the intervention of human experts.

The rest sections are organized as follows: The second section summarizes related research work. The third section introduces the representation of temporal values and time expressions in detail. The fourth section describes the framework and main components of ARTime. The fifth section reports the evaluation results of ARTime on two benchmarks. The last section concludes this paper.

\section{Related Work}
Understanding time expressions in natural language has long attracted the attention of researchers. The TIDES research program proposed TIMEX \citep{TIMEX} and TIMEX2 \citep{TIMEX2}, which are standalone annotation schemes of time expression with detailed descriptions of temporal values. The TERQAS workshops conceptualized TimeML \citep{TimeML} based on TIMEX and TIMEX2. TimeML became an ISO standard in 2009. \citet{SCATE} pointed out that the classic annotation schemes failed to show the semantic composition structure of of time expressions and proposed the Semantically Compositional Annotation of Time Expressions (SCATE). However, applying SCATE to existing corpus requires to manually re-annotate the expressions in a more complex way, and many of the existing SOTA methods can not handle annotation in SCATE format directly \citep{SemEval-2018-6}.

On the recognition of time expression, an early study shows that the complexity of time expressions is limited, and finite state automata or regex expression can be effective for recognizing those expressions \citep{FASTUS}. Mainstream recognition methods can be roughly divided to surface-structure-based methods \citep{GUTime,HTime10,HTime13,SUTime,UWTime,SynTime,PTime} and sequential-tagging-model-based methods \citep{ClearTK,CogCompTime,TOMN}. 
Research work in recent years achieves significant improvements on the recognition. SynTime \citep{SynTime} defines generic but heuristic rules on a group of time-related triggering token types. TOMN \citep{TOMN} uses the SynTime defined token types instead of the classic BIO-tagging scheme for the CRF model. PTime \citep{PTime} generalizes time expressions in training data to sequential patterns and selects a subset of the patterns for recognition. However, these studies only focus on the recognition.

The normalization of time expression is dominated by methods with expert designed rules or grammars. HeidelTime \citep{HTime10, HTime13, HTime14} uses regex rules on time tokens and modifiers to combine recognized tokens and filter ambiguous expressions. SUTime \citep{SUTime} proposes a 3-layered temporal pattern language. It firstly extends recognized tokens to string, then composes and filters the strings to get temporal values. \citet{LIDP13} use an EM-style bootstrapping approach to learn a PCFG parser on pre-defined preternminals. UWTime \citep{UWTime} uses a combinatory categorical grammar to parse possible meanings of time expressions. It selects meanings for recognized expressions via a linear classifier with context-dependent features. CogCompTime \citep{CogCompTime} provides a rule-based standalone normalizer conceptually built on \citet{CogNorm}, which achieves the SOTA normalization results on the \citep{TEval3} dataset. There are also some efforts on understanding event-related expressions. \citet{CLINICAL} analyzes time expressions in clinical notes. TweetTime \citep{TweeTime} improves existing methods by establishing an external event knowledge base. According to existing studies \citep{TEval3,TweeTime}, rule-engineering can achieve good results on covered expressions but are hard to extend to emerging corpora.

In this paper, we focus on automatically recovering the semantic structure of expressions without any compositional annotations. The latest work on recognition inspired our idea of using surface form patterns to activate normalization rules, and we replace the labor cost of designing rules by the automatic rule generation.

\begin{table*}[tbh]
\centering
\caption{The temporal operations used in ARTime.}
\label{tbl:action}
\begin{tabu} to 0.9\textwidth {X[3, l] X[10, l]}
    \toprule
    \rowfont[c]\bfseries Action & Description \\
    \midrule
    \multirow{2}{*}{\textsf{ModifyVal[$v,f$]}} & Modify the value in $f$ to $v$.\\
    & (e.g,  \textsf{ModifyVal[5,Day,Week]}(``2021-05-17'')=``2021-05-21'')\\
    \multirow{2}{*}{\textsf{ModifyEnum[$e$]}} & Use the enumerable constant $e$ to modify the corresponding field.\\
    & (e.g,  \textsf{ModifyVal[Summer]}(``2021-05-17'')=``2021-SU'')\\
    \multirow{2}{*}{\textsf{CountEnum[$v,e,f$]}} & Find the $v$-th $e$ in field $f$.\\
    & (e.g,  \textsf{CountEnum[1,Friday,Month]}(``2021-05-17'')=``2021-05-07'')\\
    \multirow{2}{*}{\textsf{Equal[$f$]}} & Let the target value equals to the base on field $f$.
    \\
    & (e.g,  \textsf{Equal[1,Friday,Month]}(``2021-05-17'')=``2021-05'')\\
    \multirow{2}{*}{\textsf{ToBegin/End[$f$]}} & Modify the value in $f$ to its begin/end point.\\
    & (e.g, \textsf{ToBegin[Month,Quarter]}(``2021-05'')=``2021-04'')\\
    \multirow{2}{*}{\textsf{For/Backward[$v,u$]}} & Increase/decrease current value by $v$ $u$. \\
    & (e.g, \textsf{Backward[2, Month]}(``2021-05'')=``2021-03'')\\
    \multirow{2}{*}{\textsf{ToNext/Last[$u$]}} & Increase/decrease current value by one $u$.\\
    & (e.g, \textsf{ToNext[Month]}(``2021-05'')=``2021-06'')\\
    \multirow{2}{*}{\textsf{MakeSet[$f$]}} & Denote that the current value are sets of $f$.\\
    & (e.g, \textsf{MakeSet[Week]}(``2021'')=``2021-WXX'')\\
    \multirow{2}{*}{\textsf{Add[$v, u$]}} & Add $v$ $u$ to the current value, only works for duration values.\\
    & (e.g, \textsf{Add[2, Month]}(``P1Y'')=``P1Y2M'')\\
    \multirow{2}{*}{\textsf{ApproxRef[$r$]}} & Mean the value is the approximate reference $r$.\\
    & (e.g, \textsf{ApproxRef[Past]}(``2021-05'')=``PAST\_REF'')\\
    \bottomrule
\end{tabu}
\end{table*}

\section{Time Expression Normalization as a Sequence of Operations
}\label{sec:def}
We model the normalization of a time expression as a sequence of \textit{operations} defined on \textit{time fields}, which can construct a \textit{temporal value} of specific type. The normalization rule is defined as a triplet consists of a \textit{surface form pattern}, a \textit{type} of temporal value, and an \textit{operation sequence}. The following subsections introduce the above concepts.

\subsection{Time Fields}
The time fields can be simply treated as time units with lower and upper bound constraints on values. Each temporal value can be denoted by a series of non-overlapping fields. For example, ISO:8601 represents a date value in the format ``yyyy-MM-DD'', where ``MM'' represents the ``month'' field with lower bound 1 and upper bound 12.\footnote{In real applications, the upper bound of a time field can denoted by a larger explicit or default time unit. For example, the field with the name ``month'' can be represented as ``monthOfYear'' or (month, year).}

\subsection{Type of Temporal Values}
According to TimeML, we classify the temporal value into 3 types according to their formats. 1) \textit{Instant} for representing date and time (e.g., ``2021-05-17T12:00''), 2) \textit{Duration} for denoting the amount of intervening time in a time interval (e.g, ``P2M'' represents 2 months.), and 3)  \textit{Approximate reference} for representing approximate referring value (e.g, ``PAST\_REF'').

\subsection{Operations}\label{sec:operation}
ARTime takes the function of time expression as changing a base temporal value to a target value. The semantic of a time expression is represented by a sequence of operations defined the temporal fields. We design ten types of operations for ARTime (as listed in Table \ref{tbl:action}. The operations take 5 kinds of parameters: 1) \textit{integer values} $v$, 2) \textit{time units} $u$, 3) \textit{temporal fields} $f$, 4) \textit{enumerable temporal constant} $e$, and 5) \textit{approximate reference} $r$ (i.e., \textsf{Past}, \textsf{Present} and \textsf{Future}).
Most of the operations are designed for temporal values of \textit{instant} type, while \textsf{ApproxRef} and \textsf{Add} are designed for \textit{approximate reference} values and \textit{duration} values respectively. Specifically, we use a \textsf{MakeSet} operation to represent the TIMEX3 type ``SET''.

In the execution of operations, we require the operations be arranged in order. Operations on larger fields should be executed first. Operations on the same fields will be arranged according to their type. The operations independent to the base (e.g, \textsf{ModifyVal}) should be executed first. The reason to use descending order of granularity is that the order corresponds to the way humans understand time fields. For example, the token ``day'' denotes ``dayOfYear'' in ``the first \underline{day} in 2021'' and ``dayOfWeek'' in ``the first \underline{day} in this week''. Its meaning depends on the larger fields mentioned in the context. Arrange operations according to their type is to prevent redundant sequences. Example \ref{eg:order} explained why executing some operations later may overriding the execution results of previous operations.

\begin{example}\label{eg:order}
Considering the base value ``2021-01'', we have
\[\begin{split}
 &\mathsf{ToNext[Month]}\left(\text{``2021-01''}\right)\\
 = & \text{``2021-02''},\\
 &\mathsf{ModifyEnum[May]}\left(\text{``2021-02''}\right)\\
 = &\mathsf{ModifyEnum[May]}\left(\text{``2021-01''}\right)\\
 = & \text{``2021-05''}
\end{split}\], which indicates that executing the subsequent \textsf{ModifyEnum[May]} might make \textsf{ToNext[Month]} a redundant operation.
\end{example}

\subsection{The Surface Form Pattern of Rule}
In our design, each rule has a surface form pattern to determine whether it can be applied to an input expression. The pattern in our approach is similar to the sequential pattern in PTime \citep{PTime}, which is defined as a sequence consisting of token types and untyped tokens. A token type consists of multiple values, and each value has a corresponding regex to capture its various surface forms. We only use 6 token types listed in Table \ref{tbl:lexicon} for obtaining variable values. The 6 types including 4 kinds of enumerable temporal constants (i.e, the first 4 rows in the table), time units, and in-equality modifiers (denoted as ``IN\_EQ'') collected from HeidelTime.

\begin{table}[hbt]
\centering
\caption{The token types.}
\label{tbl:lexicon}
\begin{tabu} to \columnwidth {X[1, c] X[2, l]}
    \toprule
    \rowfont[c]\bfseries Type & Contents \\
    \midrule
    MONTH & January, Jan., Feb., etc.\\
    WEEK & Sunday, Sun., etc,\\
    SEASON & Spring, Summer, etc.\\
    DAY\_TIME & moring, afternoon, etc.\\
    TIME\_UNIT & year, month, etc.\\
    IN\_EQ & a mere, no more than, etc.\\
    \bottomrule
\end{tabu}
\end{table}

In our method, only the tokens referring to temporal values that appear in the operation sequences will be generalized to the corresponding type. For example, the token ``day'' in rule (Pattern=``several \underline{day} later'', Type=ApproximateReference, Operations=(\textsf{ApproxRef[FUTURE\_REF]}) is not generalized to corresponding type ``TIME\_UNIT'' since the operations do not require a unit variable.

\begin{figure}[ht]
    \centering
    \includegraphics[width=0.8\columnwidth]{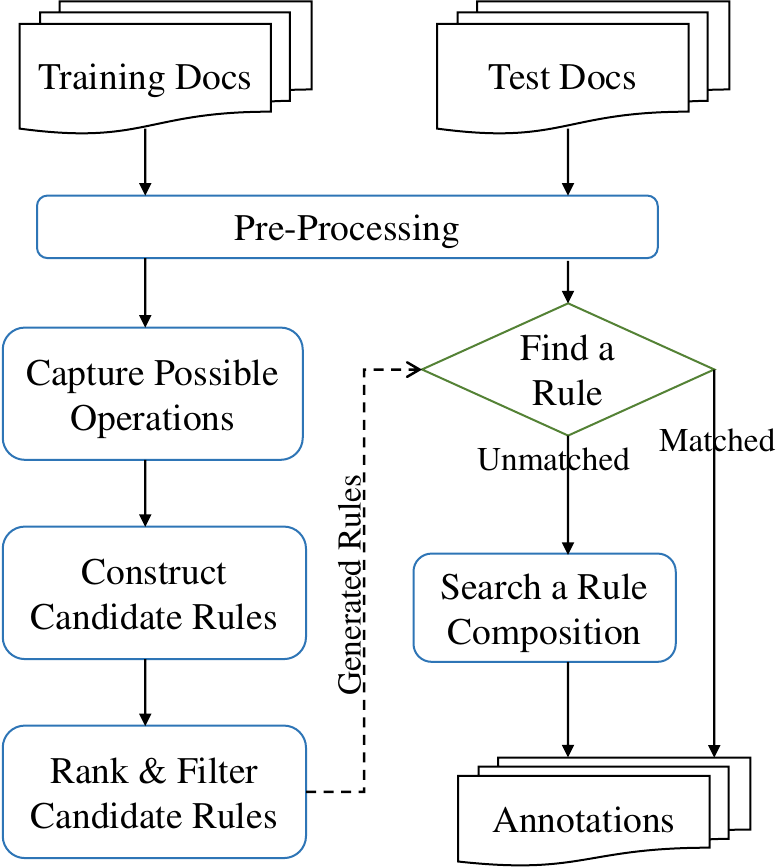}
    \caption{The framework of ARTime.}
    \label{fig:frame}
\end{figure}

\section{Framework of ARTime}
Figure \ref{fig:frame} illustrates the normalization process of ARTime. The pre-processing step is adopted from the corresponding components in PTime. The rest normalization procedures can be divided into two parts, 1) generating rules (i.e., the left part of Figure \ref{fig:frame}) and 2) applying the generated rules (i.e. the right part of Figure \ref{fig:frame}). Since the TimeML standard does not annotate the base value of each time expression, we simply use the document creation time as a substitute in capturing the possible operation sequences. The following sub-sections describe the key techniques in ARTime. Section \ref{sec:infer} details how to capture possible operation sequences. Section \ref{sec:cons} describes how to generate rules from the noisy results. Section \ref{sec:compose} describes how to use the generated rules to normalize input time expressions.

\subsection{Capturing Possible Operations}\label{sec:infer}
\begin{figure}[tbh]
    \centering
    \includegraphics[width=0.65\columnwidth]{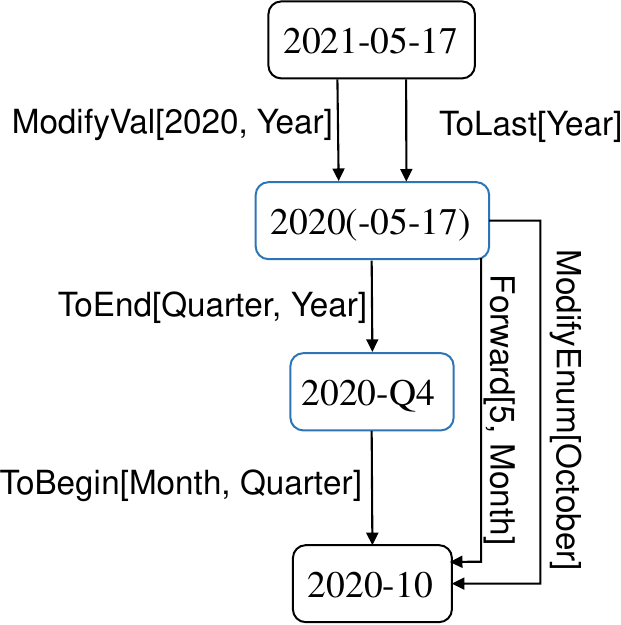}
    \caption{Some operation paths from ``2021-05-17'' to ``2020-10''}
    \label{fig:graph}
\end{figure}

By regarding temporal values as vertices and operations as directed edges connecting the base values to the normalized values, the task of reasoning possible operation sequence can be formalized as searching paths on the graph of temporal values, where each path corresponds to a sequence of operations (as demonstrated in Figure \ref{fig:graph}). The main challenge is that there could be a great quantity of paths between two values, and not all of them correspond to meaningful expression in daily communications (e.g., the sequence (\textsf{ToEnd[Quarter,Year],ToBegin[Month,Quarter]}) is legal in semantic but unnatural).

\begin{algorithm}[th]
    \caption{The DFS algorithm for changing $V_c$ to $V_t$, where $ pool$ is the pool of usable numeric values for acceleration.}
    \label{alg:smp}
    \begin{algorithmic}[1]
	\Function{DFS}{$V_c$, $V_t$, $f$, $ pool$}
	\If {$f = 1/\infty$}
	\Comment{{\small$1/\infty$ is a virtual field for the termination condition}}
		 \State \Return{$V_c = V_t$}
	\EndIf
	\State $S \leftarrow \emptyset$
	\For {$f' \in \left\{f'|1/\infty\le f' < f \right\}$}
	    \State $\Delta = V_{t[f:f']} - V_{c[f:f']}$
	    \If {$\Delta = 0$}
	        \State $S += \textsc{DFS}(V_c, V_t, f',  pool)$
	    \EndIf
	    \For {$a \subset \{\mathrm{operations\,on\,} f\}$}
	        \If {$V_c.exec(a) - V_c \ne \Delta$}
	            \State \textbf{continue}
	        \EndIf
	        \If {$\neg (\mathrm{numVals}(a) \subseteq vPool)$}
	            \State \textbf{continue}
	        \EndIf
	        \State $ pool' \leftarrow  pool - \mathrm{numVals}(a)$
	        \State $V' \leftarrow V_c.exec(a)$
	        \State $sol \leftarrow \textsc{DFS}(V', V_t, f',  pool')$
	        \If {$\text{haveSolution}$}
	            \State $S = S \cup (a + sol)$
		    \EndIf      
        \EndFor
	\EndFor
	\State \Return{S}   
	\EndFunction
    \end{algorithmic}
\end{algorithm}

Our method is based on the assumption that practical time expressions are low-redundancy sequences. i.e., we prefer direct sequences like \textsf{(Equal[Day])}(``today'') rather than the complex ones of the same meaning such as \textsf{(ToLast[Week],Forward[37,Day])}(``7 days after a week ago'').

We implement the process by a heuristic depth-first search (DFS) algorithm described in Algorithm \ref{alg:smp}. The main idea is to guide the search process by the difference between the base value and the annotated value. In each iteration, we ensure that the current value $V_c$ and the target value $V_t$ are the same on fields of granularity not less than the iterated field $f$. (line 1). We enumerate a smaller field $f'$ (line 8) and check if there are some operations $a$ on field $f'$ corresponds the difference between $V_c$ and $V_t$ from $f'$ to $f$ (line 13-14). In the enumeration of $a$ (line 12), we only consider no-redundancy sequences of the partial order introduced in section \ref{sec:operation}.  Specifically, we accelerate the process by requiring all numeric values that appear in the search results must also appear in the input time expression (line 15-16).

Given the time expression $T$ with annotated value $V_a$ and the base time $V_b$, we obtain possible operation sequences by calling $\mathrm{DFS}(V_b, V_a, \infty, \mathrm{numVals}(T))$, where $\infty$ is a virtual time unit as the initialization condition and $\mathrm{numVals}$ is the function for collecting appeared numeric values.

\subsection{Constructing and Filtering Rules}\label{sec:cons}
All the captured operation sequences will be used for constructing candidate rules. We firstly find the values appear in both the surface form and the operation sequence, then replace its appearance with corresponding token types and variable symbols to construct candidate rules. For example, given the expression ``this month'' and operation sequence \textsf{Equal[Month]}, the replacement result will be ``this TIME\_UNIT:\$1'' and \textsf{Equal[\$1]}.

The generation produces many noises since there are more than one sequence from one time value to another. We distinguish good rules by a quite simple intuition that more general patterns and more correct rules should appears on more expressions. We rank the candidate rules by their frequency and the frequency of their patterns on training corpus, then select the most frequent rules for normalization. We suppose that there is no need to drop the low frequency rules. The reason is that a low frequency rule either be replaced by more generalized rules (e.g., the second rule in Example \ref{eg:noise}), or do capture some meaningful token patterns that are difficult to generalize (e.g., ``as soon as possible'').

\begin{example}\label{eg:noise}
Consider the expression ``last month'' and normalized value ``2021-04'' and suppose that there are two candidate rules,\\
$\cdot$ (``last TIME\_UNIT:\$1'' $\Rightarrow$ \textsf{ToLast[\$1]}),\\
$\cdot$ (``last month'', $\Rightarrow$ \textsf{ModifyEnum[April]}).\\
The first one is correct and can handle similar expressions (e.g., ``last year''), while the second one only holds on the coincidence appearance of the base value ``2021-05'' .
\end{example}

\subsection{Applying Rules For Normalization}\label{sec:compose}
Given an input expression, ARTime will try to find a matchable rule to normalize it. If it can not match any generated rules, ARTime will attempt to search a consecutive composition of rules and stop words to cover it. The stop words include connecting symbols (e.g., ``-''), determiners (e.g., ``this''), prepositions (e.g., ``to'') and so on. 

The search process is performed by a segmentation algorithm (i.e., the dynamic programming algorithm described in Algorithm \ref{alg:segment}.) The algorithm tries to cover the input expression except for stop words in it (line 6-7) with minimum rules (line 11-13). For the case that there are multiple compositions of the same size, we simply choose the one that contains the most frequent rules.
After that, we assume that all the operations in chosen rules are useful and merge them into a new sequence according to the order described in section \ref{sec:operation}.

\begin{algorithm}[th]
    \caption{The segmentation algorithm for unmatched expressions.}
    \label{alg:segment}
    \begin{algorithmic}[1]
	\Function{Segment}{$T$: expression, $\mathcal{R}$: rules}
	\State Initalize $F\leftarrow$ to an array of empty sets.
	\State $F_{[0]} \leftarrow \{\emptyset\}$
	\For {$i \leftarrow 1$ to $| T |$}
	    \State $C \leftarrow \emptyset$
	    \If {$\mathrm{isStopword}( T_{[i]}) \wedge F_{[i-1]} \ne \emptyset$}
	        \State $C \leftarrow C \cup \{F_{[i-1]}\}$
	    \EndIf
	    \For {$j \leftarrow 0$ to $i - 1$}
	        \If {$F_{[{j}]} = \emptyset$}
	            \State \textbf{continue}
	        \EndIf
	        \If {$\exists r\in\mathcal{R}. \, \mathrm{match}(r,  T_{[j+1:i]})$}
                \State $C \leftarrow C \cup \{F_{[j]} \cup \{r\} \}$
	        \EndIf
	        \If {$C \ne \emptyset$}
	            \State $F_{[i]} = \mathrm{argmin}_{c\in C} |c|$
	       \EndIf
	    \EndFor
	\EndFor
	\State \Return{$F_{| T |}$}
	\EndFunction
    \end{algorithmic}
\end{algorithm}

\section{Evaluation}
\subsection{Datasets}
We use the TempEval-3 \cite{TEval3} benchmark and the Tweets benchmark proposed by \citet{SynTime}.\footnote{Benchmarks with lots of event-related time expressions like Wikiwars \citep{Wikiwars} and \citet{TweeTime}'s tweets dataset are not used in our evaluations. The reason is that understanding those expressions requires the external knowledge of the events, which is not our focus.} The statistics of the two benchmarks are illustrated in Table \ref{tbl:datasets}.

\begin{description}[leftmargin=0pt]
    \item[TempEval-3] \citep{TEval3} is a sub-task in SemEval 2013 consisting of English news articles. We follow the previous study \citep{UWTime} to use corrected TimeBank \cite{TimeBank} and AQUAINT as its training datasets.
    \item[Tweets] \citep{SynTime} is a new benchmark consisting of English tweets. The annotators tend to annotate years in a finer granularity (e.g. the annotation \textit{``... in $\langle$T value=2014-XX-XX$\rangle$2014$\langle$/T$\rangle$''} means ``a day in 2014''.) These annotations are legal according to TimeML, but do not conform to the intuition of expert designed rules in existing methods. Thus we provide the alter-version \textbf{Tweets-M} by annotating the year expressions as is.
\end{description}

\begin{table}[t]
    \centering
    \caption{The statistics of the datasets. The \textbf{Doc.}, \textbf{Token}, and \textbf{Exp.} columns report the number of documents, tokens, and time expressions in the datasets respectively.}
    \label{tbl:datasets}
    \begin{tabu} to \columnwidth {X[5, c] X[2, r] X[2, r] X[2, r]}
        \toprule
        \rowfont[c]\bfseries Dataset & Doc. & Token & Exp.\\
        \midrule 
        TimeBank & 183 & 61,418 & 1,243 \\
        AQUAINT & 73 & 33,973 & 579 \\
        TempEval-3 Eval & 20 & 6,375 & 138 \\
        Tweets train & 742 & 15,571 & 892 \\
        Tweets test & 200 & 4,198 & 237 \\
        \bottomrule
    \end{tabu}
\end{table}

\begin{table*}[t]
\centering
\caption{The accuracy(\%) of normalization results on gold recognition annotations. The best results are in \textbf{bold}, and the second-best results are \underline{underlined}.}
\label{tbl:gold}
\begin{tabu} to 0.8\textwidth {X[2, c] X[1, r] X[1, r] X[1, r] X[1, r] X[1, r] X[1, r]}
\toprule
\rowfont[c]\bfseries \multirow{2}{*}{Method} & \multicolumn{2}{c}{\textbf{TempEval-3}} & \multicolumn{2}{c}{\textbf{Tweets}} & \multicolumn{2}{c}{\textbf{Tweets-M}}\\
\cmidrule(lr){2-3} \cmidrule(lr){4-5} \cmidrule(lr){6-7}
 & Type & Value & Type & Value & Type & Value \\
\midrule 
HeidelTime & 81.2 & 76.1 & 76.4 & 66.2 & 76.4 & 71.3 \\
SUTime  & 83.3 & 70.3 & 89.5 & 83.5 & 89.5 & 88.6\\
UWTime  & 88.4 & \underline{82.6} & 76.4 & 71.3 & 76.4 & 76.4\\
CogCompN  & \textbf{91.3} & \textbf{83.4} & 86.5 & 70.9 & 86.5 & 75.9 \\
ARTime & 84.8 & 75.4 & \underline{93.2} & \textbf{87.3} & \underline{93.2} & \underline{89.0}\\
ARTime+H & \underline{90.6} & 81.9 & \textbf{94.5} & \underline{84.4} & \textbf{94.5} & \textbf{89.5} \\
\bottomrule
\end{tabu}
\end{table*}

\subsection{Compared Methods}
We compare ARTime with 4 normalization systems, \textbf{HeidelTime} \citep{HTime13}, \textbf{UWTime} \citep{UWTime}, \textbf{SUTime} \citep{SUTime} and \textbf{CogCompN} \citep{CogCompTime}. HedidelTime is the SOTA purely-rule-based system. UWTime achieves the SOTA performances on TempEval-3. SUTime outperforms the other ones on social media texts according to \citet{TweeTime}. CogCompN is the standalone normalizer of CogCompTime \citep{CogCompTime} which achieves SOTA results on TempEval-3.

\begin{table*}[t]
\centering
\caption{The end-to-end results(\%) on TempEval-3 and Tweets. The best results are in \textbf{bold}, and the second-best results are \underline{underlined}.}
\label{tbl:e2e}
\begin{tabu} to 0.95\textwidth {X[1.2, c] X[1.5, c] X[1, r] X[1, r] X[1, r] X[1, r] X[1, r] X[1, r] X[1, r] X[1, r]}
\toprule
\rowfont[c]\bfseries \multicolumn{2}{c}{\multirow{2}{*}{Method}} & \multicolumn{4}{c}{TempEval-3} & \multicolumn{4}{c}{Tweets-M}\\
\cmidrule(lr){3-6} \cmidrule(lr){7-10}
& & Type & \multicolumn{3}{c}{Value} & Type & \multicolumn{3}{c}{Value}\\
\cmidrule(lr){1-2} \cmidrule(lr){3-3} \cmidrule(lr){4-6} \cmidrule(lr){7-7} \cmidrule(lr){8-10}
Reco. & Norm. & F1 & Pr & Re & F1 & F1 & Pr & Re & F1 \\
\midrule
\multicolumn{2}{c}{HeidelTime} & 83.3 & 80.2 & 76.1 & 78.1 & 84.4 & 88.0 & 71.3 & 78.8\\
\multicolumn{2}{c}{SUTime} & 81.9 & 67.8 & 70.3 & 69.0 & 87.8 & 85.4 & \underline{88.6} & 87.0 \\
\multicolumn{2}{c}{UWTime} & 85.7 & \textbf{85.9} & 79.7 & \textbf{82.7} & 83.6 & \textbf{93.7} & 74.7 & 83.1 \\
\midrule
\multirow{3}{*}{SynTime} & CogCompN & 88.5 & 80.0 & \textbf{81.2} & 80.6 & 86.5 & 77.0 & 74.7 & 75.8\\
 & ARTime &  86.3 & 78.6 & 74.6 & 76.6 & 93.9 & \underline{91.9} & 86.5 & 89.1\\
 & ARTime+H & \textbf{90.1} & 82.2 & \underline{80.4} & \underline{81.3} & 94.4 & 90.3 & 86.5 & 88.4\\
\midrule
\multirow{3}{*}{TOMN} & CogCompN & \underline{89.3} & 82.0 & 79.0 & 80.4 & 86.1 & 75.7 & 73.4 & 74.5\\
 & ARTime & 86.2 & 80.3 & 71.0 & 75.4 & 89.3 & 91.1 & 86.1 & 88.5\\
 & ARTime+H & 88.7 & \underline{82.8} & 76.8 & 79.7 & 93.3 & 89.5 & 86.1 & 87.7\\
\midrule
\multirow{3}{*}{PTime}
& CogCompN & 85.5 & 82.4 & 78.3 & 80.3 & 88.0 & 76.7 & 76.4 &76.5\\
& ARTime & 83.0 & 75.8 & 72.5 & 74.1 & \underline{94.7} & 89.7 & \underline{88.6} & \underline{89.2}\\
 & ARTime+H & 86.0 & 79.9 & 77.5 & 78.7 & \textbf{95.2} & 89.1 & \textbf{89.5} & \textbf{89.3}\\
\bottomrule
\end{tabu}
\end{table*}

We also evaluate the performance of compared normalization methods in real applications. We implement end-to-end systems with 3 SOTA recognition methods, SynTime \citep{SynTime}, TOMN \citep{TOMN}, and PTime \citep{PTime}. We directly use the output of HeidelTime, SUTime, and UWTime for end-to-end comparison because that they use the same rules (or grammar) for recognition and normalization.

\subsection{Evaluation Metrics}
We use the scripts\footnote{\url{https://bitbucket.org/kentonl/uwtime/src/master/evaluation_tools/}} provided by TempEval-3 for evaluation. For the normalization results, we report the accuracy of normalized temporal results with gold mentions. For the end-to-end results, we report the F1 score of normalized types, and the precision (Pr), recall (Re), and F1 score of normalized temporal values. 

\subsection{Experimental Results}
\subsubsection{Normalization Results}
Table \ref{tbl:gold} reports the normalization results on gold recognition annotations. ARTime surpasses other methods and shows better adaptability and robustness on Tweets (i.e., +3.8 points on the original Tweets).
The performances of the compared methods dramatically vary on the different corpus. All compared methods except SUTime achieve very poor results on Tweets, while SUTime achieves the worst results on TempEval-3. ARTime's performances are not very well on TempEval-3. The main reason is that the training data and the test data of TempEval-3 are annotated separately, and the insufficiency of training data severely hurts the performance of purely data-driven methods like ARTime according to previous study \cite{PTime}. For example, the test data of TempEval-3 includes 2 expressions about ``flu season’’ (It should be normalized as winter), our method cannot handle them since none of the training expressions contains the word ``season’’. Besides, the normalized values of some expressions rely on the tenses of corresponding utterances and need to be re-computed by post-modification \cite{HTime10,UWTime}. (e.g, ``finished in \underline{June}'' denotes ``June in last year'' for base temporal values like ``2021-05''). The above problems can be alleviated by introducing prior knowledge. We transform the expert rules in HeidelTime into ARTime's formats as pre-defined rules, name the combined approach as ARTime+H. ARTime+H achieves a good balance on different domains with the best results on Tweets-M and competitive results on TempEval-3. (i.e, 1.5 points lower than the SOTA results on values.)

\subsubsection{End-to-end Results}
Table \ref{tbl:e2e} reports the end-to-end results on TempEval-3 and Tweets-M. ARTime with the SOTA recognition method (PTime) outperforms the existing methods with an improvement of +2.2 points on the F1 scores of normalized values on Tweets-M. The results of ARTime on TempEval-3 are not good enough, but can be easily improved by introducing the same prior knowledge used in HeidelTime. ARTime+H with SynTime achieve the second-best results on the F1 score on values without losing the advantages on Tweets (1.4 points higher than the best results achieved by compared methods).

\subsubsection{Analysis}
We categorize the negative samples in the normalization results of ARTime by their causes in Table \ref{tbl:error}. About half of the negative samples are due to unseen patterns that can not be captured by our rules. Another problem is the errors caused by tense in the context. Some existing systems apply post-modification tricks by comparing the tense to the positivity of the difference between the output value and the base value. If our method can correctly utilize the oracle tense information, the accuracy on TempEval-3 can increase to 79.7\% (+4.3 points). There are also some cases that the rules generated in our method do not fit the input expressions (The 3rd row in Table \ref{tbl:error}).

\begin{table}[t]
\centering
\caption{The statistics(\%) of negative samples in the normalization results}
\label{tbl:error}
\begin{tabu} to \columnwidth {X[2.5, l] X[2, r] X[1.6, r]}
    \toprule
    \rowfont[c]\bfseries Errors & TempEval-3 & Tweets-M \\
    \midrule
    Unseen Pattern & 41.2 & 50.0\\
    Tense Error & 17.6 & 11.5\\
    Bad Rule & 8.8 & 19.2\\
    Annotation Error & 8.8 & 3.8\\
    Others & 23.5 & 15.4\\
    \bottomrule
\end{tabu}
\end{table}

We also manually analyzed the rules used in the test process to show what extent the introduction of expert rules replaces the automatic generation in ARTime+H, the results are illustrated in Table \ref{tbl:rnum}. The ``Full'' column reports the number of rules used in normalizing the expressions, and the ``Auto'' and ``Ratio'' columns report how many of those rules can be covered by automatically generation. From the results we can know that the automatic generation can cover over 90 percent of the manual rules and adding about 2 rules are enough for ARTime.

\begin{table}[t]
\centering
\caption{The statistics of rules in the normalization results of ARTime+H.}
\label{tbl:rnum}
\begin{tabu} to \columnwidth {X[4, l] X[2, r] X[2, r] X[3, r]}
    \toprule
    \rowfont[c]\bfseries Dataset & Auto & Full & Ratio(\%)\\
    \midrule
    TempEval-3 & 34 & 36 & 91.9\%\\
    Tweets-M & 40 & 42 & 95.2\%\\
    \bottomrule
\end{tabu}
\end{table}

\subsubsection{Running Efficiency}
All the results of ARTime are obtained by a single-threaded Scala implementation on a personal workstation with an Intel Xeon CPU E5-1607 v4 @ 3.10GHz CPU and 128GB RAM. In average, ARTime generates $\sim$4.8 candidate rules for each expression. The offline training process took $\sim$16.3 minutes on TempEval-3 and $\sim$13.5 minutes on Tweets. The test process took $\sim$47 seconds on TempEval-3 and $\sim$46 seconds on Tweets.

\section{Conclusion}
In this paper, we mainly focus on automatically generating rules for time expression normalization. The main contributions of this paper are summarized as follows:

$\cdot$ We model time expression normalization as an operation sequence to construct the normalized temporal value, and ten basic operations are defined for time expression normalization. 

$\cdot$ We present a novel method, called ARTime, for generating normalization rules from training data without expert interventions. Specifically, ARTime captures possible operation sequences from annotated data and generates candidate rules on time expressions with common surface forms, and finally obtains normalization rules by ranking the candidate rules. 

$\cdot$ Our experimental results show that ARTime outperforms SOTA methods on the Tweets benchmark, and achieves competitive results with existing expert-engineered rule methods on the Tempeval-3 benchmark. The end-to-end results when combining ARTime with time expression recognition systems are also very competitive.

There are still some rooms to improve ARTime. One of the future work is to generate more high-quality rules. The other is to enable ARTime to use the tense and event information in context.

\section*{Acknowledgements}
This work is supported by the National Science Foundation of China under grant No.61772264. We would like to thank our team members Guanji Gao and Yanjia Wang for their help in the early exploration stage of this work.

\bibliography{custom}

\begin{thebibliography}{24}
\expandafter\ifx\csname natexlab\endcsname\relax\def\natexlab#1{#1}\fi

\bibitem[{Angeli and Uszkoreit(2013)}]{LIDP13}
Gabor Angeli and Jakob Uszkoreit. 2013.
\newblock \href {https://www.aclweb.org/anthology/P13-1009}
  {Language-independent discriminative parsing of temporal expressions}.
\newblock In \emph{Proceedings of the 51st Annual Meeting of the Association
  for Computational Linguistics (Volume 1: Long Papers)}, pages 83--92, Sofia,
  Bulgaria. Association for Computational Linguistics.

\bibitem[{Bethard(2013)}]{ClearTK}
Steven Bethard. 2013.
\newblock \href {https://www.aclweb.org/anthology/S13-2002}
  {{C}lear{TK}-{T}ime{ML}: A minimalist approach to {T}emp{E}val 2013}.
\newblock In \emph{Proceedings of the Seventh International Workshop on
  Semantic Evaluation}, pages 10--14, Atlanta, Georgia, USA. Association for
  Computational Linguistics.

\bibitem[{Bethard and Parker(2016)}]{SCATE}
Steven Bethard and Jonathan Parker. 2016.
\newblock A semantically compositional annotation scheme for time
  normalization.
\newblock In \emph{Proceedings of the Tenth International Conference on
  Language Resources and Evaluation}, pages 3779--3786.

\bibitem[{Chang and Manning(2012)}]{SUTime}
Angel~X. Chang and Christopher Manning. 2012.
\newblock \href
  {http://www.lrec-conf.org/proceedings/lrec2012/summaries/284.html} {Sutime: A
  library for recognizing and normalizing time expressions}.
\newblock In \emph{Proceedings of the Eight International Conference on
  Language Resources and Evaluation}, pages 3735--3740, Istanbul, Turkey.
  European Language Resources Association.

\bibitem[{Ding et~al.(2019)Ding, Gao, Shi, and Qu}]{PTime}
Wentao Ding, Guanji Gao, Linfeng Shi, and Yuzhong Qu. 2019.
\newblock \href {https://doi.org/10.1609/aaai.v33i01.33016335} {A pattern-based
  approach to recognizing time expressions}.
\newblock In \emph{The Thirty-Third {AAAI} Conference on Artificial
  Intelligence}, pages 6335--6342, Honolulu, Hawaii, USA. {AAAI} Press.

\bibitem[{Ferro et~al.(2005)Ferro, Gerber, Mani, Sundheim, and Wilson}]{TIMEX2}
Lisa Ferro, Laurie Gerber, Inderjeet Mani, Beth Sundheim, and George Wilson.
  2005.
\newblock \href
  {https://www.ldc.upenn.edu/sites/www.ldc.upenn.edu/files/english-timex2-guidelines-v0.1.pdf}
  {Standard for the annotation of temporal expressions-tides}.
\newblock \emph{The MITRE Corporation, McLean-VG-USA}.

\bibitem[{Hobbs et~al.(1997)Hobbs, Appelt, Bear, Israel, Kameyama, Stickel, and
  Tyson}]{FASTUS}
Jerry~R Hobbs, Douglas Appelt, John Bear, David Israel, Megumi Kameyama, Mark
  Stickel, and Mabry Tyson. 1997.
\newblock \href {http://arxiv.org/abs/cmp-lg/9705013} {Fastus: A cascaded
  finite-state transducer for extracting information from natural-language
  text}.
\newblock \emph{arXiv preprint cmp-lg/9705013}.

\bibitem[{Laparra et~al.(2018)Laparra, Xu, Elsayed, Bethard, and
  Palmer}]{SemEval-2018-6}
Egoitz Laparra, Dongfang Xu, Ahmed Elsayed, Steven Bethard, and Martha Palmer.
  2018.
\newblock {S}em{E}val 2018 task 6: Parsing time normalizations.
\newblock In \emph{Proceedings of The Twelfth International Workshop on
  Semantic Evaluation}, pages 88--96, New Orleans, Louisiana. Association for
  Computational Linguistics.

\bibitem[{Lee et~al.(2014)Lee, Artzi, Dodge, and Zettlemoyer}]{UWTime}
Kenton Lee, Yoav Artzi, Jesse Dodge, and Luke Zettlemoyer. 2014.
\newblock \href {https://doi.org/10.3115/v1/P14-1135} {Context-dependent
  semantic parsing for time expressions}.
\newblock In \emph{Proceedings of the 52nd Annual Meeting of the Association
  for Computational Linguistics (Volume 1: Long Papers)}, pages 1437--1447,
  Baltimore, Maryland. Association for Computational Linguistics.

\bibitem[{Mazur and Dale(2010)}]{Wikiwars}
Pawel Mazur and Robert Dale. 2010.
\newblock \href {https://www.aclweb.org/anthology/D10-1089} {{W}iki{W}ars: A
  new corpus for research on temporal expressions}.
\newblock In \emph{Proceedings of the 2010 Conference on Empirical Methods in
  Natural Language Processing}, pages 913--922, Cambridge, MA. Association for
  Computational Linguistics.

\bibitem[{Ning et~al.(2018)Ning, Zhou, Feng, Peng, and Roth}]{CogCompTime}
Qiang Ning, Ben Zhou, Zhili Feng, Haoruo Peng, and Dan Roth. 2018.
\newblock \href {https://doi.org/10.18653/v1/D18-2013} {{C}og{C}omp{T}ime: A
  tool for understanding time in natural language}.
\newblock In \emph{Proceedings of the 2018 Conference on Empirical Methods in
  Natural Language Processing: System Demonstrations}, pages 72--77, Brussels,
  Belgium. Association for Computational Linguistics.

\bibitem[{Pustejovsky et~al.(2003)Pustejovsky, Hanks, Sauri, See, Gaizauskas,
  Setzer, Radev, Sundheim, Day, Ferro et~al.}]{TimeBank}
James Pustejovsky, Patrick Hanks, Roser Sauri, Andrew See, Robert Gaizauskas,
  Andrea Setzer, Dragomir Radev, Beth Sundheim, David Day, Lisa Ferro, et~al.
  2003.
\newblock \href
  {http://ucrel.lancs.ac.uk/publications/cl2003/papers/pustejovsky.pdf} {The
  timebank corpus}.
\newblock In \emph{Corpus linguistics}, volume 2003, page~40. Lancaster, UK.

\bibitem[{Pustejovsky et~al.(2010)Pustejovsky, Lee, Bunt, and Romary}]{TimeML}
James Pustejovsky, Kiyong Lee, Harry Bunt, and Laurent Romary. 2010.
\newblock \href
  {http://www.lrec-conf.org/proceedings/lrec2010/pdf/55_Paper.pdf}
  {{ISO}-{T}ime{ML}: An international standard for semantic annotation}.
\newblock In \emph{Proceedings of the Seventh International Conference on
  Language Resources and Evaluation}, pages 394--397, Valletta, Malta. European
  Language Resources Association.

\bibitem[{Setzer and Gaizauskas(2000)}]{TIMEX}
Andrea Setzer and Robert Gaizauskas. 2000.
\newblock \href {http://www.lrec-conf.org/proceedings/lrec2000/pdf/321.pdf}
  {Annotating events and temporal information in newswire texts}.
\newblock In \emph{Proceedings of the Second International Conference on
  Language Resources and Evaluation}, Athens, Greece. European Language
  Resources Association.

\bibitem[{Str{\"o}tgen et~al.(2014)Str{\"o}tgen, B{\"o}gel, Zell, Armiti, Canh,
  and Gertz}]{HTime14}
Jannik Str{\"o}tgen, Thomas B{\"o}gel, Julian Zell, Ayser Armiti, Tran~Van
  Canh, and Michael Gertz. 2014.
\newblock \href
  {http://www.lrec-conf.org/proceedings/lrec2014/pdf/849_Paper.pdf} {Extending
  {H}eidel{T}ime for temporal expressions referring to historic dates}.
\newblock In \emph{Proceedings of the Ninth International Conference on
  Language Resources and Evaluation}, pages 2390--2397, Reykjavik, Iceland.
  European Language Resources Association.

\bibitem[{Str{\"o}tgen and Gertz(2010)}]{HTime10}
Jannik Str{\"o}tgen and Michael Gertz. 2010.
\newblock \href {https://www.aclweb.org/anthology/S10-1071} {{H}eidel{T}ime:
  High quality rule-based extraction and normalization of temporal
  expressions}.
\newblock In \emph{Proceedings of the 5th International Workshop on Semantic
  Evaluation}, pages 321--324, Uppsala, Sweden. Association for Computational
  Linguistics.

\bibitem[{Str{\"o}tgen et~al.(2013)Str{\"o}tgen, Zell, and Gertz}]{HTime13}
Jannik Str{\"o}tgen, Julian Zell, and Michael Gertz. 2013.
\newblock \href {https://www.aclweb.org/anthology/S13-2003} {{H}eidel{T}ime:
  Tuning {E}nglish and developing {S}panish resources for {T}emp{E}val-3}.
\newblock In \emph{Proceedings of the Seventh International Workshop on
  Semantic Evaluation}, pages 15--19, Atlanta, Georgia, USA. Association for
  Computational Linguistics.

\bibitem[{Tabassum et~al.(2016)Tabassum, Ritter, and Xu}]{TweeTime}
Jeniya Tabassum, Alan Ritter, and Wei Xu. 2016.
\newblock \href {https://doi.org/10.18653/v1/D16-1030} {{T}wee{T}ime : A
  minimally supervised method for recognizing and normalizing time expressions
  in {T}witter}.
\newblock In \emph{Proceedings of the 2016 Conference on Empirical Methods in
  Natural Language Processing}, pages 307--318, Austin, Texas. Association for
  Computational Linguistics.

\bibitem[{Tissot et~al.(2015)Tissot, Roberts, Derczynski, Gorrell, and
  Del~Fabro}]{CLINICAL}
Hegler Tissot, Angus Roberts, Leon Derczynski, Genevieve Gorrell, and
  Marcus~Didonet Del~Fabro. 2015.
\newblock \href {https://www.aclweb.org/anthology/W15-0211} {Analysis of
  temporal expressions annotated in clinical notes}.
\newblock In \emph{Proceedings of the 11th Joint {ACL}-{ISO} Workshop on
  Interoperable Semantic Annotation}, London, UK. Association for Computational
  Linguistics.

\bibitem[{UzZaman et~al.(2013)UzZaman, Llorens, Derczynski, Allen, Verhagen,
  and Pustejovsky}]{TEval3}
Naushad UzZaman, Hector Llorens, Leon Derczynski, James Allen, Marc Verhagen,
  and James Pustejovsky. 2013.
\newblock \href {https://www.aclweb.org/anthology/S13-2001} {{S}em{E}val-2013
  task 1: {T}emp{E}val-3: Evaluating time expressions, events, and temporal
  relations}.
\newblock In \emph{Proceedings of the Seventh International Workshop on
  Semantic Evaluation}, pages 1--9, Atlanta, Georgia, USA. Association for
  Computational Linguistics.

\bibitem[{Verhagen et~al.(2005)Verhagen, Mani, Sauri, Littman, Knippen, Jang,
  Rumshisky, Phillips, and Pustejovsky}]{GUTime}
Marc Verhagen, Inderjeet Mani, Roser Sauri, Jessica Littman, Robert Knippen,
  Seok~B. Jang, Anna Rumshisky, John Phillips, and James Pustejovsky. 2005.
\newblock \href {https://doi.org/10.3115/1225753.1225774} {Automating temporal
  annotation with {TARSQI}}.
\newblock In \emph{Proceedings of the {ACL} Interactive Poster and
  Demonstration Sessions}, pages 81--84, Ann Arbor, Michigan. Association for
  Computational Linguistics.

\bibitem[{Zhao et~al.(2012)Zhao, Do, and Roth}]{CogNorm}
Ran Zhao, Quang Do, and Dan Roth. 2012.
\newblock \href {https://www.aclweb.org/anthology/N12-3008} {A robust shallow
  temporal reasoning system}.
\newblock In \emph{Proceedings of the Demonstration Session at the Conference
  of the North {A}merican Chapter of the Association for Computational
  Linguistics: Human Language Technologies}, pages 29--32, Montr{\'e}al,
  Canada. Association for Computational Linguistics.

\bibitem[{Zhong and Cambria(2018)}]{TOMN}
Xiaoshi Zhong and Erik Cambria. 2018.
\newblock \href {https://doi.org/10.1145/3178876.3185997} {Time expression
  recognition using a constituent-based tagging scheme}.
\newblock In \emph{Proceedings of the 2018 World Wide Web Conference}, page
  983–992, Republic and Canton of Geneva, CHE. International World Wide Web
  Conferences Steering Committee.

\bibitem[{Zhong et~al.(2017)Zhong, Sun, and Cambria}]{SynTime}
Xiaoshi Zhong, Aixin Sun, and Erik Cambria. 2017.
\newblock \href {https://doi.org/10.18653/v1/P17-1039} {Time expression
  analysis and recognition using syntactic token types and general heuristic
  rules}.
\newblock In \emph{Proceedings of the 55th Annual Meeting of the Association
  for Computational Linguistics (Volume 1: Long Papers)}, pages 420--429,
  Vancouver, Canada. Association for Computational Linguistics.

\end{thebibliography}
\bibliographystyle{acl_natbib}

\end{document}